\documentclass[journal]{IEEEtran}
\ifCLASSINFOpdf
\else
\fi
\usepackage{CJK,CJKnumb}
\usepackage{subfigure}
\usepackage{graphicx}
\usepackage{bm}
\usepackage{indentfirst}
\usepackage{float}
\usepackage[cmex10]{amsmath}
\begin{document}
%
\title{Vision-aided Localization and Navigation Based on Trifocal Tensor}
%
%
%

\author{Qiang~Fang  
\thanks{Q. Fang is with the College of Mechatronics and Automation, National University of Defense Technology, Hunan, Changsha, China. e-mail: (qiangfang@nudt.edu.cn,qiang-fang@hotmail.com)}}
\maketitle

\begin{abstract}
  In this paper, a novel method for vision-aided navigation based on trifocal tensor is presented. The main goal of the proposed method is to provide position estimation in GPS-denied environments for vehicles equipped with a standard inertial navigation systems(INS) and a single camera only. We treat the trifocal tensor as the measurement model, being only concerned about the vehicle state and do not estimate the the position of the tracked landmarks. The performance of the proposed method is demonstrated using simulation and experimental data.
\end{abstract}


%
\IEEEpeerreviewmaketitle

\section{Introduction}
%
%
%
%
Accurate navigation state estimation is essential in many fields, and Global Positioning System(GPS) aided inertial measurement unit(IMU) navigation is the primary method today. However, GPS might be unavailable or unreliable, such as operating indoors, under water or on other planets. In these scenarios, vision-based methods constitute an attractive alternative for navigation aiding due to their relatively low cost and autonomous nature.\par
Vision-aided navigation$^{[1-2]}$ has indeed become an active research field over the past decades. Many works have addressed visual-inertial motion estimation, and several camera-IMU solutions have been proposed to track the state of a system in real-time on computationally constrained platforms and in real environments. Existing methods vary by the number of overlapping images and by the techniques used for fusing the imagery data with the navigation system. Considering two overlapping images, it is only possible to determine camera rotation and up-to-scale translation$^{[3]}$(translation during the interval is associated with an unknown scale). Therefore, two-view based methods for navigation aiding$^{[4-5]}$ are incapable of eliminating the developing navigation errors in all states.\par
Given multiple images($>$2), it is possible to determine the camera motion up to a common scale$^{[3]}$(translations during the intervals are associated with a common unknown scale). Considering the computational requirements, several muti-view methods using the augmented state technique for navigation aiding have been already proposed$^{[6-9]}$, an approach commonly referred to as simultaneous localization and mapping (SLAM)$^{[10-13]}$. For the ability to handle loops(vehicle returns to the original region), [14] proposed to cope with loops using bundle adjustment$^{[15]}$, however, this method might be hardly possible in the real-time performance.\par
In contrast to SLAM, [16] presented a method based on three-view geometry. Compared with bundle adjustment and SLAM, the proposed method reduces the computational resources. Through linearizing the residual measurement and calculating the relevant Jacobian matrix, the authors obtained good position results by applying the implicit extended Kalman filter(IEKF). In order to provide better position estimations, [17] proposed an unscented Kalman filter(UKF) method based on the idea of three-view geometry. Given three overlapping images, the familiar constraints are the trifocal tensor$^{[3]}$, and in 1997, Hartley$^{[18]}$ applied the trifocal tensor to solve the main problem of computing the motion of the camera (along with the scene structure). For uncalibrated image sequences, Torr$^{[19]}$ focused on the problem of degeneracy in structure and motion recovery using trifocal tensor. Guerrero$^{[20]}$  directly solved the trifocal tensor using singular value decomposition (SVD) method for location and mapping. Based on the augmented state technique$^{[7]}$, Hu$^{[21]}$ added the trifocal tensor to the multi-view measurement model to estimate the current vehicle state and the features' state during some interval.\par
In this paper, we present an algorithm that is able to fuse the  INS and vision based on trifocal tensor. Our approach is motivated by the three-view geometry constraints derived by Indelman$^{[16]}$, which is different from the augmented state technique. We treat the trifocal tensor as the measurement model, being only concerned about the vehicle state, which means the position of the tracked landmarks does not have to be estimated.  It is obvious that the constraints of the trifocal tensor and  three-view geometry are both derived from three overlapping images, so there might be some connection between them. After a brief discussion of the reference frames in the next section, the trifocal tensor will be described and its relationship with the three-view geometry analysed in section III; in section IV the details of the IEKF estimator based on trifocal tensor will be presented; the performance of the algorithm will be verified by simulation and experimental results in section V; and finally, in section VI the conclusions of this work will be drawn.
\section{Reference frames}
In this section the reference frames used to derive the Vision/INS equations are summarized as follows:\par
1)	$n$--North-East-Down(NED) coordinate system. Its origin is set at the location of the navigation system.  It points North,  East, and completes a Cartesian right hand system.\par
2)	$b$--Body-fixed reference frame. Its origin is set at center-of-mass of the vehicle.  It points towards front of the vehicle, or right when viewed from above, and  completes the setup to yield a Cartesian right hand system.\par
3)	$c$--Camera-fixed reference frame. Its origin is set at the center-of-projection of the camera.  It points toward the field-of-view(FOV) center, or toward the right half of the FOV when viewed from the center-of-projection of the camera, and completes the setup to yield a Cartesian right hand system.

\section{Trifocal Tensor}
We first briefly describe the trifocal tensor and then analyze its relationship with the three-view geometry developed by Indelman ${[16]}$.\par
Figure 1 shows a 3-view point correspondence, where $ \bm X$ is a space point,  $\bm{\widehat{x},\widehat{x}^{'},\widehat{x}^{''}}$ are the space point $\bm X$ projected in the first, second and third camera image respectively, $ c_1,c_2,c_3 $ are the camera centres of the first, second and third camera respectively, and $ \pi,\pi^{'},\pi^{''}$ are the image planes.
\begin{figure}[h]
 \centering
  \includegraphics[width=0.45\textwidth]{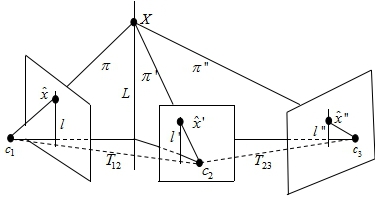}
 \caption{3-view point correspondence among three camera images}\label{fig:digit}
\end{figure}

Assuming the camera calibration matrix is known, then the camera projection matrices can be defined as$^{[3]}$
\begin{equation}
 \bm{U}=[\bm{I}|\bm{0}],\bm{U}^{'}=[\bm{A}|\bm{a}_{4}],\bm{U}^{''}=[\bm{B}|\bm{b}_{4}]
\end{equation}
where $ \bm{A}=[\bm{a}_{1},\bm{a}_{2},\bm{a}_{3}]=\bm{C}_{c_{1}}^{c_{2}}$ is the rotation matrix from camera $c_{1}$ to camera $c_{2}$, $ \bm{B}=[\bm{b}_{1},\bm{b}_{2},\bm{b}_{3}]=\bm{C}_{c_{1}}^{c_{3}}$ is the rotation matrix from camera $c_{1}$ to camera $c_{3}$,$\bm  a_{4}=\bm{T}_{12}^{c_{2}} $ is the translation expressed in camera  $c_{2}$ from $c_{1}$ camera to camera $c_{2}$, and $\bm  b_{4}=\bm{T}_{13}^{c_{3}} $ is the translation expressed in camera $c_{3}$ from camera $c_{1}$ to camera $c_{3}$.\par
    According to [3], the trifocal tensor incidence relations using matrix notation can be given as
\begin{equation}
[\widehat{\bm{x}}^{'}\times](\sum_{i=1}^{3}\widehat{x}_{i}\bm{T}_{i})[\widehat{\bm{x}}^{''}\times]=\bm{0}_{3\times3}
\end{equation}
where $ \bm{T}_{i}=\bm{a}_{i}\bm{b}_{4}^{T}-\bm{a}_{4}\bm{b}_{i}^{T}(i=1,2,3)$ and image points are normalized as the LOS vectors expressed in the camera systems, that is, $ \widehat{\bm{x}}=[\widehat{x}_{1},\widehat{x}_{2},\widehat{x}_{3}]^{T}=[x_{1},y_{1},f_{1}]^{T},\widehat{\bm{x}}^{'}=[\widehat{x}^{'}_{1},\widehat{x}^{'}_{2},\widehat{x}^{'}_{3}]^{T}=[x_{2},y_{2},f_{2}]^{T},\widehat{\bm{x}}^{''}=[\widehat{x}^{''}_{1},\widehat{x}^{''}_{2},\widehat{x}^{''}_{3}]^{T}=[x_{3},y_{3},f_{3}]^{T}$, $ x_{i},y_{i}$ is the point expressed in the camera system $c_{i}$ and $f_{i}$ is the camera focal length. \par
   The variables used in eq.(2) are expressed in camera systems. Rearranging eq.(2), we can yield the trifocal tensor expression in the navigation system as(the proof is in the Appendix A):
 \begin{equation}
   [\bm{\widehat{x}}^{'n}\times]\bm{\widehat{x}}^{n}\bm{T}_{23}^{nT}[\bm{\widehat{x}}^{''n}\times]-[\bm{\widehat{x}}^{'n}\times][(\bm{\widehat{x}}^{n}\times\bm{T}_{12}^{n})\times][\bm{\widehat{x}}^{''n}\times]=\bm{0}_{3\times3}
\end{equation}
where $\bm{\widehat{x}}^{n}=\bm{C}_{c_{2}}^{n}\bm{\widehat{x}}^{'},\bm{\widehat{x}}^{'n}=\bm{C}_{c_{2}}^{n}\bm{\widehat{x}}^{'},\bm{\widehat{x}}^{''n}=\bm{C}_{c_{2}}^{n}\bm{\widehat{x}}^{''}$,$\bm{T}_{23}^{n}=\bm{Pos}_{3}^{n}-\bm{Pos}_{2}^{n}$ and $\bm{T}_{12}^{n}=\bm{Pos}_{2}^{n}-\bm{Pos}_{1}^{n}$. Here $\bm{C}_{c_{i}}^{n}=\bm{C}_{b_{i}}^{n}(\bm{\theta})\bm{C}_{c_{i}}^{b_{i}}(i=1,2,3)$ is the rotation matrix from camera $ c_{i}$ to the navigation system( $\bm{C}_{b_{i}}^{n}$ is the rotation matrix from the body system at time $t_{i}$ to the navigation system which is changed with the vehicle's rotation angle $ \bm{\theta}$ and $\bm{C}_{c_{i}}^{b_{i}}$ is the rotation matrix from camera $c_{i}$ to  body system $ b_{i}$, which is calibrated in the beginning and maintained unchanged), and $\bm{Pos}_{i}^{n}(i=1,2,3)$ is the position expressed in the navigation system at time $ t_{i}$.\par
As we introduced in section I, the constraints of the trifocal tensor and the three-view geometry are both derived from three overlapping images. There should be some connection between them. Through the analysis, we have obtained the following Lemma(the detailed proof is listed in the Appendix B):\par
\emph{Lemma 1}:  The trifocal tensor constraints are the sufficient conditions of the three-view geometry constraints.\par
 According to [16] , given multiple matching features, one can determine the translation vectors $\bm{T}_{12}$ and $\bm{T}_{23}$, respectively, up to scale. In general, these two scale unknowns are different. The two scales can be connected through three-view geometry, which relates between the magnitudes of $\bm{T}_{12}$ and $\bm{T}_{23}$. Consequently, if the magnitude of $\bm{T}_{12}$ is known, it is possible to calculate both the direction and the magnitude of $\bm{T}_{23}$, and in [16], the authors have obtained good results. Then from Lemma 1, we might conclude that the performance of the trifocal tensor constraints is better than that of the three-view geometry used in a navigation system such as [16]. And this conclusion will be verified in the following sections.
 
\section{Estimator description}
\subsection{Structure of the EKF state vector}
The evolving INS state is described by the vector$^{[7]} $
\begin{equation}
\bm{X}=[\bm{q}_{n}^{bT},\bm{b}_{g}^{T},\bm{V}_{n}^{T},\bm{b}_{a}^{T},\bm{Pos}_{n}^{T}]^{T}
\end{equation}
where $\bm{q}_{n}^{b} $ is the unit quaternion describing the rotation from frame n to b, $\bm{Pos}_{n} $ and $\bm{V}_{n} $ are the vehicle position and velocity with respect to frame n, and finally $ \bm{b}_{g}$
and $ \bm{b}_{a}$ are $ 3\times1$  vectors that describe the biases affecting the gyroscope and accelerometer measurements, respectively. The IMU biases are modeled as random walk processes,
driven by the white Gaussian noise vectors $\bm{n}_{wg}$ and $\bm{n}_{wa}$, respectively. Following eq.(4), the IMU error-state is defined as:
\begin{equation}
\delta\bm{X}=[\delta\bm{\theta}^{T},\delta\bm{b}_{g}^{T},\delta\bm{V}_{n}^{T},\delta\bm{b}_{a}^{T},\delta\bm{Pos}_{n}^{T}]^{T}
\end{equation}

For the position, velocity, and biases, the error is defined as the standard additive error, i.e., the estimate $\widehat{x}$ of a quantity $ x$ is defined as $ \delta{x}=x-\widehat{x}$. However,for
the quaternion,the error quaternion $ \delta{q}$ describes the small error rotation, and it is defined as
\begin{equation}
  \delta{\bm{q}}=[0.5\delta\bm{\theta}^{T}   1]^{T}
\end{equation}
where $\delta\bm{\theta}$ is the attitude error.
\subsection{System model}
The linearized continuous-time model for the IMU error-state is $^{[7]} $
\begin{equation}
\delta\dot{\bm{X}}=\bm{F}\delta\bm{X}+\bm{G}\bm{n}_{IMU}
\end{equation}
where $ \bm{n}_{IMU}=[\bm{n}_{g}^{T},\bm{n}_{wg}^{T},\bm{n}_{a}^{T},\bm{n}_{wa}^{T}]$ is the system
noise. The matrices $\bm{F} $ and $\bm{G} $ are \[
 \bm{F}=
\begin{bmatrix}
-[\widehat{\bm{w}}\times] &-\bm{I_{3}} &\bm{0}_{3\times3} &\bm{0}_{3\times3} &\bm{0}_{3\times3} \\
\bm{0}_{3\times3} &\bm{0}_{3\times3} &\bm{0}_{3\times3} &\bm{0}_{3\times3} &\bm{0}_{3\times3}\\
-\bm{C}_{b}^{n}[\widehat{\bm{a}}\times] &\bm{0}_{3\times3} &-2[\bm{w}_{G}\times] &-\bm{C}_{b}^{n} &-[\bm{w}_{G}\times]^{2}\\
\bm{0}_{3\times3} &\bm{0}_{3\times3} &\bm{0}_{3\times3} &\bm{0}_{3\times3} &\bm{0}_{3\times3}\\
\bm{0}_{3\times3} &\bm{0}_{3\times3} &\bm{I_{3}} &\bm{0}_{3\times3} &\bm{0}_{3\times3}\\
\end{bmatrix}
\]
where $ \bm{I_{3}}$ is the $3\times3 $  identity matrix,$\bm{w}_{G}$ is the planet's rotation,
$\widehat{\bm{w}}=\bm{w}_{m}-\widehat{\bm{b}}_{g}-\bm{C}_{n}^{b}\bm{w}_{G}$ and $\widehat{\bm{a}}=\bm{a}_{m}-\widehat{\bm{b}}_{a} $,$ \bm{w}_{m}$ is the gyroscope measurements  and $ \bm{a}_{m}$ is the accelerometer measurements,and
\[
\bm{G}=
\begin{bmatrix}
-\bm{I}_{3} &\bm{0}_{3\times3} &\bm{0}_{3\times3} &\bm{0}_{3\times3}\\
\bm{0}_{3\times3} &\bm{I}_{3} &\bm{0}_{3\times3} &\bm{0}_{3\times3}\\
\bm{0}_{3\times3} &\bm{0}_{3\times3} &-\bm{C}_{b}^{n} &\bm{0}_{3\times3}\\
\bm{0}_{3\times3} &\bm{0}_{3\times3} &\bm{0}_{3\times3} &\bm{I}_{3}\\
\bm{0}_{3\times3} &\bm{0}_{3\times3} &\bm{0}_{3\times3} &\bm{0}_{3\times3}
\end{bmatrix}
\]

\subsection{Mesurement model}
Define matrix $\bm{M}=[\bm{\widehat{x}}^{'n}\times]\bm{\widehat{x}}^{n}\bm{T}_{23}^{nT}[\bm{\widehat{x}}^{''n}\times]-[\bm{\widehat{x}}^{'n}\times][(\bm{\widehat{x}}^{n}\times\bm{T}_{12}^{n})\times][\bm{\widehat{x}}^{''n}\times] $,and the elements of matrix  $\bm{M}$ are
 \begin{equation}
 \bm{M}=
  \begin{bmatrix}
 m_{11} & m_{12} & m_{13}\\
 m_{21} & m_{22} & m_{23}\\
 m_{31} & m_{32} & m_{33}\\
 \end{bmatrix}\\
 \end{equation}
where $ m_{ij}$ represents the row $i$ and column $j$ element of $ \bm{M}$, and \\
\begin{equation}
\begin{aligned}
m_{ij}=&-(\bm{\widehat{x}}^{'n}\times\bm{\widehat{x}}^{nT})\bm{E}_{ij}(\bm{\widehat{x}}^{''n}\times\bm{T}_{23}^{n})-(\bm{\widehat{x}}^{'n}\times\bm{\widehat{x}}^{n})^{T}\bm{E}_{ij}(\bm{\widehat{x}}^{''n}\\
&\times\bm{T}_{12}^{n})+(\bm{\widehat{x}}^{'n}\times\bm{T}_{12}^{n})^{nT}\bm{E}_{ij}(\bm{\widehat{x}}^{''n}\times\bm{\widehat{x}}^{n})
 \end{aligned}
\end{equation}
with $\bm{E}_{ij}$ representing a $3\times3 $ dimension matrix, and the row $i$ and column $j$ element being one and others zeros.\par
Noting that the trifocal tensor constraints in eq.(3) contain nine trilinearites, only four are linearly independent$^{[3]}$. Geometrically these four trilinearites arise from special choices of the lines in the second and third images for the point-line-line relation. \par
Here we choose $m_{11},m_{13},m_{31},m_{33} $ as the basis, which means we choose a line parallel to the image x-axis and a line through the image coordinate origin in the second and third images respectively. Then the trifocal tensor constraints in eq.(3) can be simplied as four linearly independent constraints.\par
 First consider a single feature, which is observed in three images captured at time instances $t_{1},t_{2},t_{3}$($t_{1}<t_{2}<t_{3}$), and then the trifocal tensor constraints can be given as follows:
 \begin{equation}
  \begin{aligned}
 \bm{z}_{1}&=[m_{11},m_{13},m_{31},m_{33}]^{T}\\
&=h_{1}(\bm{Pos}_{t_{1}},\bm{Pos}_{t_{2}},\bm{Pos}_{t_{3}},\bm{\theta}_{t_{1}},\bm{\theta}_{t_{2}},\bm{\theta}_{t_{3}},\bm{\widehat{x}},\bm{\widehat{x}^{'}},\bm{\widehat{x}}^{''})
 \end{aligned}
\end{equation} \par
While in typical scenarios, the number of the matching pairs of features is usually not only one, assuming there are $ N$ matching features from all the three views. Then the measurement model can be written as
 \begin{equation}
\bm{z}=h(\bm{Pos}_{t_{1}},\bm{Pos}_{t_{2}},\bm{Pos}_{t_{3}},\bm{\theta}_{t_{1}},\bm{\theta}_{t_{2}},\bm{\theta}_{t_{3}},\bm{\widehat{x}},\bm{\widehat{x}^{'}},\bm{\widehat{x}}^{''})
=\begin{pmatrix}
\bm{z}_{1}\\
\vdots\\
\bm{z}_{N}
\end{pmatrix}
\end{equation} \par

\subsection{Implicit Extended Kalman Filtering}
In this section, we present the implicit Extended Kalman Filter(IEKF) used to analyze the performance of the fusion with a navigation system.\par
 Equation (11) shows that the measurement model $\bm{z}$ implicitly contains the system states(position $\bm{Pos}$, attitude $\bm{\theta}$)and the measurement noise of the images. Linearizing $\bm{z}$ about $\bm{Pos}_{t_{1}},\bm{\theta}_{t_{1}},\bm{\theta}_{t_{2}},\bm{Pos}_{t_{2}},\bm{Pos}_{t_{3}},\bm{\theta}_{t_{3}}$, and $\bm{\widehat{x}},\bm{\widehat{x}^{'}},\bm{\widehat{x}}^{''}$, and keeping the first-order yields
\begin{equation}
\bm{z}\approx\bm{H_{3}}\delta\bm{X_{t_{3}}}+\bm{H_{2}}\delta\bm{X_{t_{2}}}+\bm{H_{1}}\delta\bm{X_{t_{1}}}+\bm{Dv}
\end{equation}
where $\bm{H_{3}},\bm{H_{2}},\bm{H_{3}}$ are defined as
\begin{equation}
\begin{aligned}
&\bm{H_{3}}=[\frac{\partial z}{\partial \bm{\theta}_{t_3{}}},\bm{0}_{3\times12},\frac{\partial \bm{z}}{\partial \bm{Pos}_{t_3{}}}]\\
&\bm{H_{2}} =[\frac{\partial z}{\partial \bm{\theta}_{t_2{}}},\bm{0}_{3\times12},\frac{\partial \bm{z}}{\partial \bm{Pos}_{t_2{}}}]\\
&\bm{H_{1}} =[\frac{\partial z}{\partial \bm{\theta}_{t_1{}}},\bm{0}_{3\times12},\frac{\partial \bm{z}}{\partial \bm{Pos}_{t_1{}}}]
\end{aligned}
\end{equation}\par
$\bm{v}$ is the image noise associated with the LOS
vectors $\bm{\widehat{x},\widehat{x}^{'},\widehat{x}^{''}}$, and its covariance matrix is R. The matrix $D$ in (12) is the gradient of $h$ with respect to the LOS vectors  and
\begin{equation}
\bm{D}=[\frac{\partial \bm{z}}{\partial \bm{\widehat{x}}},\frac{\partial \bm{z}}{\partial \bm{\widehat{x}^{'}}},\frac{\partial \bm{z}}{\partial \bm{\widehat{x}^{''}}}]\\
\end{equation}\par
During the time interval $(t_{k},t_{k+T})$, the IMU state estimate $\bm{X}^{-} $is propagated using
 Runge-Kutta numerical integration of the basic INS equations in [7]. Moreover, the propagation step of the filter is carried out
using state transition matrix $\bm{\Phi}$ computed by numerical integration of the differential equation
\begin{equation}
\dot{\bm{\Phi}}(t_{k+\tau},t_{k})=\bm{F}\bm{\Phi}(t_{k+\tau},t_{k}),\tau\in(0,T)
\end{equation} \par
The propagation covariance is similarly calculated by numerical integration of the Lyapunov equation:
\begin{equation}
\dot{\bm{P}}=\bm{F}\bm{P}+\bm{P}\bm{F}+\bm{G}\bm{Q}_{IMU}\bm{G}^{T}
\end{equation}
where $ \bm{Q}_{IMU}$ is the covariance matrix of system noise $ \bm{n}_{IMU}$.\par
Noting that we are only interested in estimating
the navigation errors at the current time instant $\bm{X}_{t_{3}} $,
the navigation errors at the first two time instances are
considered as random parameters in the measurement
equation. Therefore, since $\bm{X}_{t_{2}} $ and $\bm{X}_{t_{1}} $ are not
estimated, these errors are transmitted in the filter
covariance matrices $\bm{P}_{t_{2}},\bm{P}_{t_{1}}$, respectively, which
are attached to the first two images.\par
   Then the Kalman gain matrix can be given by$^{[16]}$
 \begin{equation}
  \bm{K}=\bm{P}_{\bm{X}_{t_{3}}\bm{z}}\bm{P}_{\bm{z}}^{-1}
 \end{equation}\par
Since $ \bm{z}^{-}=\bm{H_{3}}\delta{\bm{X}}^{-}_{t_{3}}$
\begin{equation}
\begin{aligned}
\delta{\bm{z}}&=\bm{z}-\bm{z}^{-}\\
&=\bm{H_{3}}\delta{\bm{X}}^{-}_{t_{3}}+\bm{H_{2}}\delta\bm{X_{t_{2}}}+\bm{H_{1}}\delta\bm{X_{t_{1}}}+\bm{Dv}
\end{aligned}
\end{equation}\par
Hence
\begin{equation}
\begin{aligned}
\bm{P}_{\bm{X}_{t_{3}}}&=\bm{P}_{3}^{-}\bm{H}_{3}^{T}+\bm{P}_{32}^{-}\bm{H}_{2}^{T}+\bm{P}_{31}^{-}\bm{H}_{1}^{T}\\
\bm{P}_{\bm{z}}&=\bm{H}_{3}\bm{P}_{3}^{-}\bm{H}_{3}^{T}+
\begin{bmatrix}
\bm{H}_{2} &\bm{H}_{1}
\end{bmatrix}
\begin{bmatrix}
\bm{P}_{2} &\bm{P}_{21}\\
\bm{P}_{21}^{T} &\bm{P}_{1}
\end{bmatrix}
\begin{bmatrix}
\bm{H}_{2} &\bm{H}_{1}
\end{bmatrix}^{T}\\&+\bm{D}\bm{R}\bm{D}^{T}
\end{aligned}
\end{equation}\par
Then the corrected state and covariance are given as
\begin{equation}
\begin{aligned}
\delta{\bm{X}}_{t_{3}}&=\bm{K}\delta{\bm{z}}\\
\bm{P}_{3}^{+}&=\bm{P}_{3}^{-}-\bm{K}\bm{P}_{\bm{z}}\bm{K}^{T}
\end{aligned}
\end{equation}\par
The estimated state $\delta{\bm{X}}_{t_{3}}$ is then used for correcting the navigation solution and IMU bias paremetrization $\bm{X}_{t_{3}}$.\par
Referring to (20), the matrices $ \bm{P}_{1},\bm{P}_{2},\bm{P}_{3}^{-}$ are known, and one of the challenges during IEKF is how to calculate the cross-correlation matrices $ \bm{P}_{21},\bm{P}_{31}^{-},\bm{P}_{32}^{-}$ . Here the term $ \bm{P}_{21}$ may be calculated as
   \begin{equation}
   \bm{P}_{21}=\bm{\Phi}(t_{2},t_{1})\bm{P}_{1}
   \end{equation}\par
    The other two cross-correlation terms, $ \bm{P}_{31}^{-},\bm{P}_{32}^{-}$, may be neglected (e.g. loops). In case the above assumptions regarding $ \bm{P}_{21},\bm{P}_{31}^{-},\bm{P}_{32}^{-}$  are not satisfied, these terms may be calculated using the methods developed in [22-24].
\subsection{Computation Requirements}
 It is interesting to examine the computational complexity of the operations needed during the IEKF, and analyze its relationship with other methods, such as the "classical" MSCKF$^{[7]}$.\par
 Assume that at some step $k $ there are $N $ features included in the three overlapping images. The computational complexity of carrying out the algorithm IEKF at step k involves the computation of the predicted state $\bm{X}^{-} $ and propagation covariance $\bm{P}_{3}^{-}$, which also requires the computation of the corresponding jacobians $\bm{F}_{k},\bm{G}_{k} $ and the updated $\bm{X}_{k} ,\bm{P}_{k}$, which involves the computation of the corresponding jacobian $\bm{H}_{i},\bm{D} $, the Kalman gain matrix $K $, the innovation $\delta{\bm{z}} $ , and its covariance $\bm{P}_{z}$. \par
Note that during the IEKF process we do not estimate the state of the features, and the state dimension r remains unchanged. Here the state dimension $r=15 $, so the computation of the predicted state $\bm{X}^{-} $ and propagation covariance $\bm{P}_{3}^{-}$ is $O(1)$. The computation of the Jacobian matrices $\bm{H}_{1},\bm{H}_{2},\bm{H}_{3},\bm{D}$ is linear in $N $, that is, $O(N)$ operations.  Consider as an example the innovation covariance matrix $\bm{P}_{z}$ in eq.(19). Normally, the computation of this matrix would require multiplications $(4Nr^{2}+16rN^{2})+(16Nr^{2}+32rN^{2})+(36N+48N^{2})$ and $(4Nr(r-1)+16(r-1)N^{2})+(8Nr(2r-1)+16(2r-1)N^{2})+(24N+32N^{2})$ sums, that is, $O(N^{2})$ operations. Similar analysis leads to the conclusion that the cost of computing the innovation $\delta{\bm{z}} $ is $O(N)$, the covariance matrix $\bm{P}_{k}$ is $O(N^{2})$, and that the greatest cost in an IEKF update is the computation of the Kalman gain matrix  $K $, which is  $O(N^{3})$. Thus, the computational cost per step of IEKF is third power of the size of the features.\par
   To the "classical" MSCKF, the IMU poses at the times the last 2 images were recorded, so the state dimension is $r=27$, and the dimension of the measurement state is $2N$. Similarly considering the innovation covariance matrix, the computation of this matrix would require multiplications $2Nr^{2}+4rN^{2}$ and $(2Nr(r-1)+4(r-1)N^{2})+(4N^{2})$ sums, that is, $O(N^{2})$ operations.
Similar analysis leads to the conclusion that the greatest cost in an MSCKF update is the computation of the Kalman gain matrix  $K $, which is $O(N^{3})$ . It is the same as that in an IEKF update.  
\section{Simulation and experimental Results} 

In this section, we use the IEKF to analyze the performance of the fusion with
a navigation system through simulation and experimental results.
\subsection{Simulation results}

\begin{table}[h]
\centering  
\caption{Initial navigation errors and IMU errors}
\begin{tabular}{cccc}  
\hline
States &Description &Value &Units\\ \hline  
$\delta\bm{Pos}$ &position(1$\delta$) &$[0.1,0.1,0.1]^{T}$ &m\\         
$\delta\bm{V}$ &velocity(1$\delta$) &$[0.1,0.1,0.1]^{T}$ &m/s\\        
$\delta\bm{\theta}$ &attitude(1$\delta$) &$[1,1,1]^{T}$ &deg\\
$\delta\bm{b}_{g}$ &drift(1$\delta$) &$[0.2,0.2,0.2]^{T}$ &deg/s\\
$\delta\bm{b}_{a}$ &drift(1$\delta$) &$[10,10,10]^{T}$ &mg\\ \hline
\end{tabular}
\end{table}\par
 We present statistical results obtained by applying the trifocal tensor constraints to a trajectory containing a loop based on a simulated navigation system and synthetic imagery features. The assumed initial navigation errors and IMU errors are summarized in Table I. The IMU is sampled at 100Hz and the synthetic imagery features are sampled at 1Hz, which are obtained by assuming camera calibration matrix $\bm{K}_{cam}$
 \[
 \bm{K}_{cam}=
  \begin{bmatrix}
 260 &0 & 376\\
 0 & 260 & 240\\
 0 &0 & 1\\
 \end{bmatrix}\\
 \]\par
 and the image noise covariance  $\bm{R}$ is
\[
 \bm{R}=
  \begin{bmatrix}
 1 &0 & 0\\
 0 & 1 & 0\\
 0 &0 & 0\\
 \end{bmatrix}\\
\]

Assume the simulated trajectory is a loop repeated once as shown in Fig.2. In order to demonstrate the performance of the algorithm in loop scenarios, two different update modes are demonstrated: 1) ``sequential update", in which all the three images are acquired closely to each other, and 2) ``loop update", in which the first two images are captured while the platform passes a given region for the first time, whereas the third image is obtained at the second passing of the same region. The total running time is about 215s and at $t=210s$ the vehicle returns to its original area, which makes it possible to deal with the loop update mode using the method developed in this paper. \par

\begin{figure}[h]
 \centering
  \begin{minipage}[b]{0.24\textwidth}
    \includegraphics[width=1\textwidth]{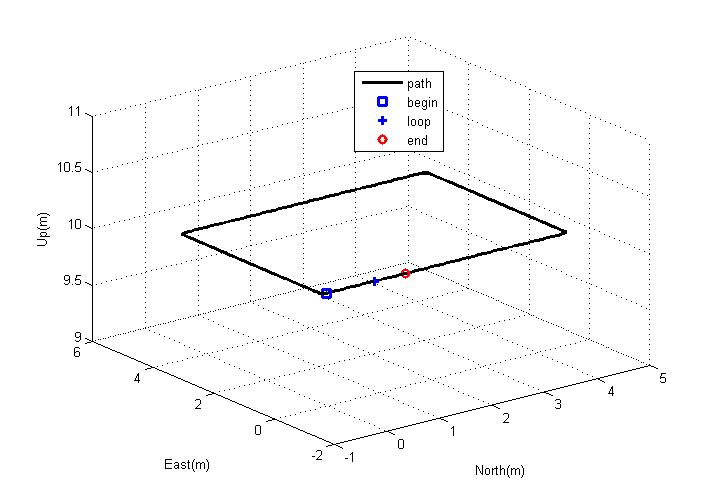}
    \caption{loop path \newline}
    \label{fig:digit}
  \end{minipage}
  \begin{minipage}[b]{0.24\textwidth}
    \includegraphics[width=1\textwidth]{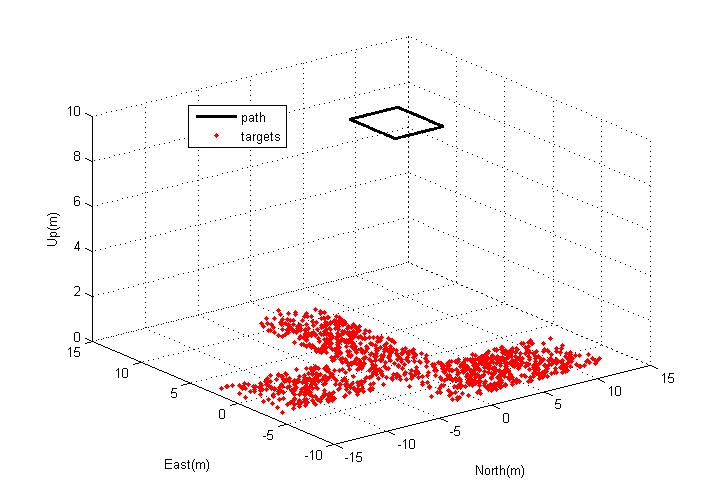}
    \caption{loop path and features}
    \label{fig:digit}
  \end{minipage}
\end{figure}\par

\begin{figure}[h]
   \centering
    \includegraphics[height=5cm,width=0.45\textwidth]{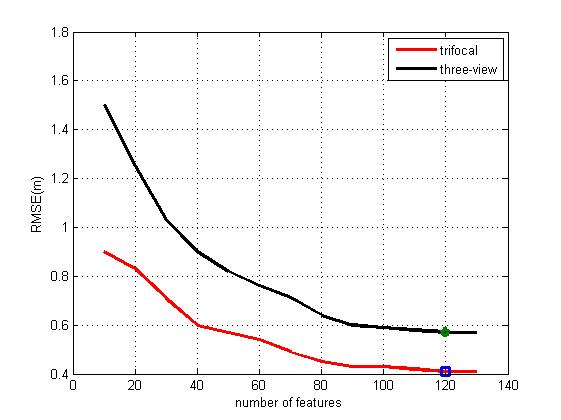}
  \caption{Navigation performance associated with features' number}
    \label{fig:digit}
  \end{figure}

During the simulation, the image features are generated randomly(Fig.3), which remain in the field of the camera view among the three overlapping images according to the flight data using the pinhole camera model, so it is necessary to analyze the impact of the features' number on the navigation performance, especially the position estimations. \par
Figure 4 shows the Monte-Carlo results(200 runs) of the RMSE(root-mean-square error) of the  position at time $t_{3}=210s$ using the developed trifocal tensor and the three-view geometry constraints.
It can be seen from Fig.4 that, as the number of the features is greater than 120, the performance of the change is
not very obvious, so in our simulation, we randomly select 120 matching triplets for the three images and use the trifocal tensor and three-view geometry constraints to analyse the system performance.\par

Figure 5 shows the Monte-Carlo results (200 runs) using the developed trifocal tensor. The curve is the square root of the filter covariance, defined for the $ith$ component in the state vector as $\sqrt{P(i,i)}$ , which is the a-posteriori covariance matrix. In addition, the three-view geometry and INS-only estimations are shown for comparison. For $t\leq150s$, the sequential upadate mode is used, the time instances $(t_{1},t_{2},t_{3})$ are chosen such that $ t_{1}=0s,t_{2}=1s$ and $t_{3}=10s$. The position and velocity errors in the INS method grow unbounded during the flight. The position errors are reset in all axes to a few decimeters(Fig.5(a)) in the trifocal tensor and three-view geometry, and the position errors of the trifocal tensor are much closer to the initial errors(0.1m). The velocity errors and accelerometer bias are considerably reduced(nearly zeros) in all axes as a result of the algorithm activation(see Fig.5(b-c)) both in the two methods. The attitude errors(Fig.5(d)) decrease to about 0.2 degrees in all axes in the trifocal tensor, while they are about 0.5 degrees in the three-view geometry. To the gyro drift(Fig.5(e)), they decrease to about 0.002 degrees per second in all axes in the trifocal tensor, however, they are about 0.007 degrees per second in the three-view geometry. Once $t=210s$, the vehicle returns to its orignal area, and it is possible to apply the algorithm in the loop update mode.
 \begin{figure}[H]
\centering
\subfigure[]{
\includegraphics[height=4cm,width=0.45\textwidth]{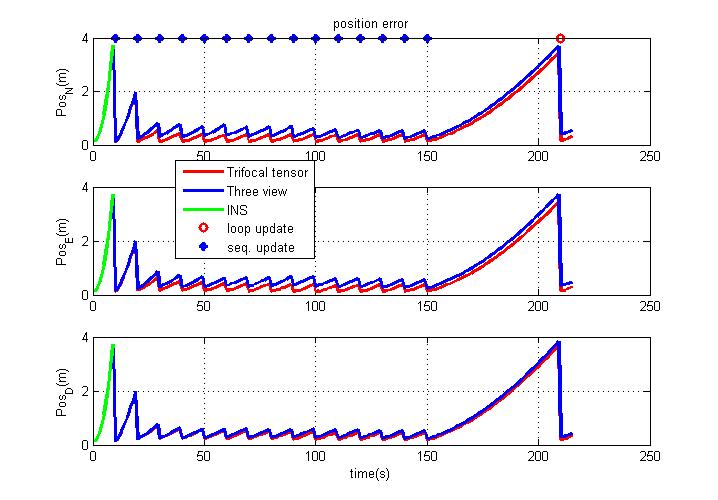}
}
\subfigure[]
{
\includegraphics[height=4cm,width=0.45\textwidth]{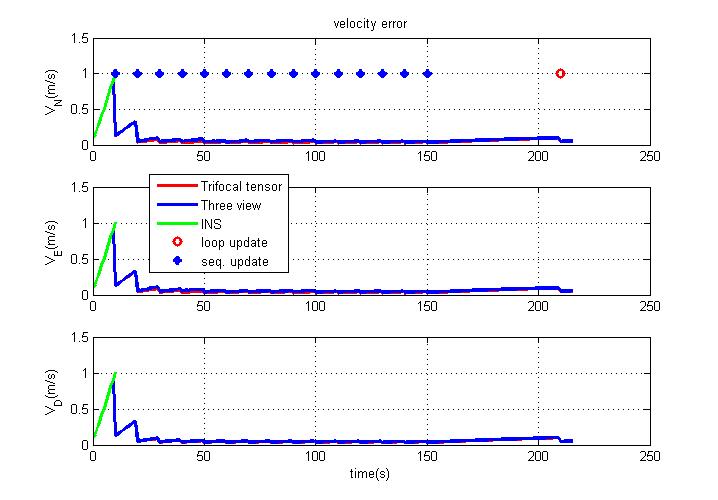}
}
\subfigure[]
{
\includegraphics[height=4cm,width=0.45\textwidth]{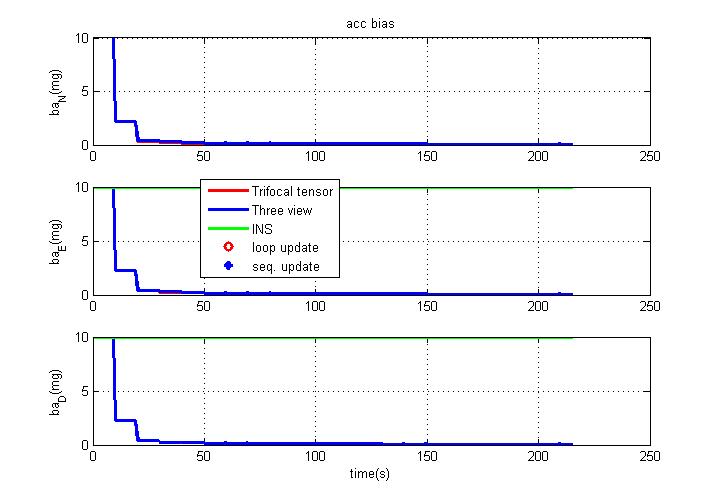}

}
\subfigure[]
{
\includegraphics[height=4cm,width=0.45\textwidth]{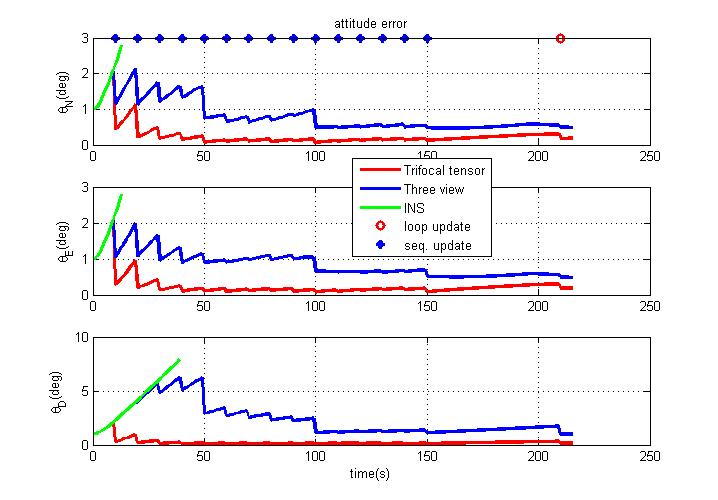}
}
\subfigure[]
{
\includegraphics[height=4cm,width=0.45\textwidth]{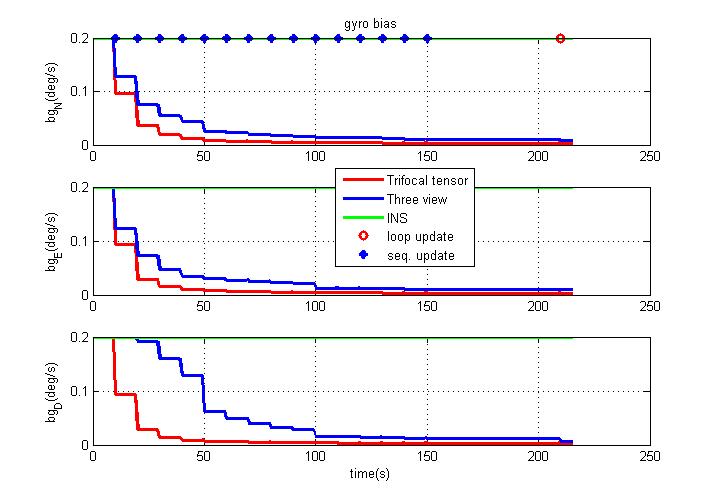}
}
\caption{200 Monte-Carlo results error estimations(covariance) using the trifocal tensor,three-view geometry and INS. (a) Position errors. (b) Velocity errors. (c)Accelerate bias (d)Attitude errors  (e)Gyro bias.} \label{fig:4}
\end{figure}
  The time instances $(t_{1},t_{2},t_{3})$ are chosen such that $ t_{1}=0s,t_{2}=1s$ and $t_{3}=210s$. Taking the position errors as an example, they are reset in all axes to a few decimeters(Fig.5(a)) from 4 meters and the position errors of the trifocal tensor are much closer to the initial errors(0.1m) than those of the three-view geometry. Similar analysis leads to the conclusion that the performance of the trifocal tensor is better than that of the three-view geometry and INS.

\subsection{Experimental results}
An experiment is carried out in this section. During the experiment, the dataset package collected by Lee$^{[25]}$ at ETHz is applied to validate the proposed method. There are five synchronized datasets in the package. Here we select the 1LoopDown dataset(the flight trajectory is a loop repeated once using the downward looking camera). The inertial sensor measurements and camera images are recorded for postprocessing at 200 Hz and 50 Hz, respectively, and the ture position of the quadrotor is obtained by the vicon system located in the room for comparison. The features are detected and matched using the SIFT${^[26]}$ algorithm added RANSAC$^{[27]}$ method.\par
\begin{figure}[h]
\centering
\subfigure[]{
\includegraphics[width=0.45\textwidth,height=0.15\textheight]{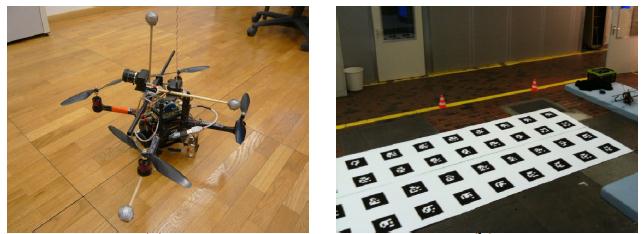}
}
\subfigure[]{
\includegraphics[width=0.45\textwidth,height=0.20\textheight]{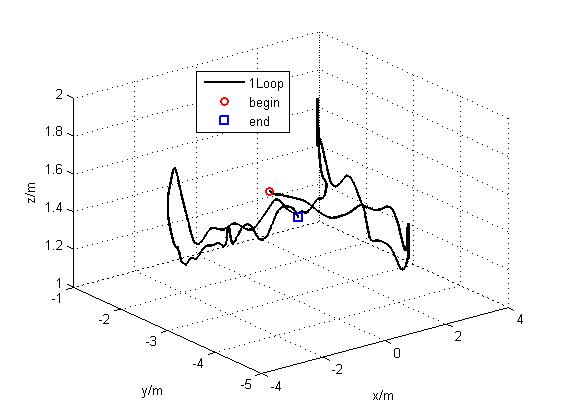}
}
\caption{Experiment processing.(a) the quadrotor and flight environment (b) flight trajectory .} \label{fig:digit}
\end{figure}\par
Figure 6(a) shows the quadrotor collecting the dataset package in the indoor flight environment and (b) shows the flight trajectory. The quadrotor flies for about 32 seconds and returns to its original area at ${t=27s}$. In the experiment, the time instances $(t_{1},t_{2},t_{3})$ are chosen such that $ t_{2}-t_{1}=0.1s$ and $t_{3}-t_{1}=1s$, and the first time instances are  $t_{1}=0s,t_{2}=0.1s,t_{3}=1s$. At first, we utilize the sequential update mode, and once the quadrotor returns to its original area, e.g., at ${t=27s}$, we begin to apply the loop update mode, and the time instances are $t_{1}=0s,t_{2}=0.1s,t_{3}=27s$. \par
Figure 7 shows the detected matching features during the experiment. It can be seen that sometimes there are no
matching features, such as at $t=2s,3s,14s,21s,22s$. In this case, we might use the INS-only update method instead, that is, we just execute the propagation step of the filter using eq.(16).


\begin{figure}[h]
 \centering
    \includegraphics[width=0.45\textwidth]{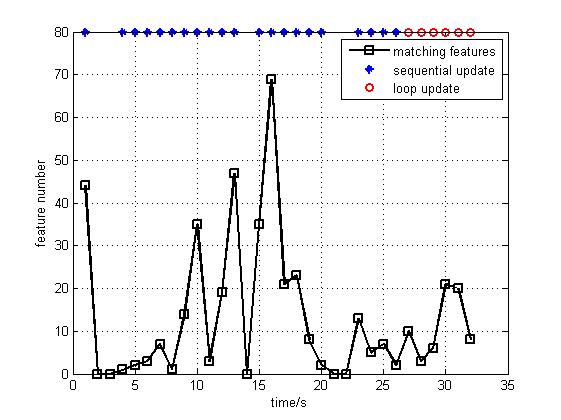}
    \caption{Number of detected features \newline}
    \label{fig:digit}
 \end{figure}\par
 
 \begin{figure}[h]
\centering
\subfigure[]{
\includegraphics[width=0.45\textwidth,height=0.22\textheight]{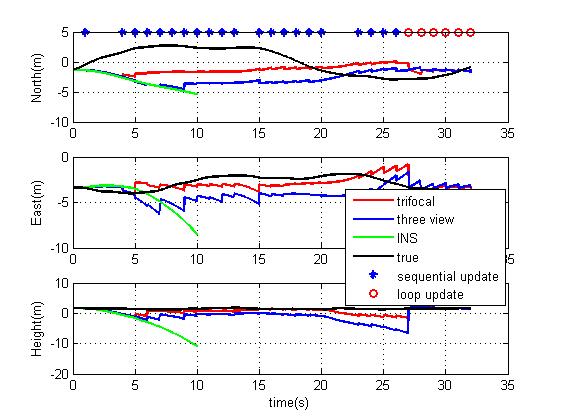}
}
\subfigure[]{
\includegraphics[width=0.45\textwidth,height=0.22\textheight]{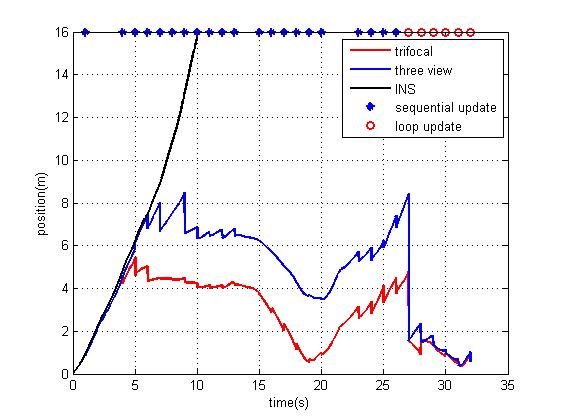}
}
    \caption{Estimated position using different methods. (a)position errors in three axes (b) the total position error  }
    \label{fig:digit}
\end{figure}\par

The position estimations are shown in Fig.8. It can be intuitively seen that the position errors increase greatly in the INS-only method, but reset in all axes to meters(Fig.8(a)) in the trifocal tensor and three-view geometry both. At time ${t=27s}$, the position errors decrease most(Fig.8(b)), because at time ${t=27s}$, the loop update mode is used. Moreover, the postion error using the trifocal tensor is much closer to zero than that of the three-view geometry.

As shown in Table II, we can also quantitatively analyze the performance of the proposed method in the mean position error, max position error and end position error. Taking the mean position error as an example, the error in the trifocal tensor method is 2.85m, which is just a half of that in the three-view geometry method and one-sixteenth of that in the INS-only method. Similar analysis on max position error and end position error leads to the conclusion that the trifocal tensor method in this paper outperforms the three-view geometry and the INS-only methods.
\begin{table}[h]
\centering  
\caption{The position errors}
\begin{tabular}{cccc}  
\hline
Algorithm  &mean position  &Max position  &End position\\
  & error(m) &error(m) &error(m)\\ \hline  
trifocal tensor &2.8500 &5.4342 &0.5801\\         
three-view &4.7225 &8.5094 &0.5944\\        
INS only  &44.9512 &120.6157 &120.6157\\
\hline
\end{tabular}
\end{table}\par

\section{Conclusion and future work}

This paper presented a new method for the purpose of Vision/INS applications based on trifocal tensor. The trifocal tensor constraints derived here were the sufficient conditions of the three-view geometry constraints. The proposed method utilized three overlapping images to formulate constraints relating between the platform motions at the time instances of the three images, and the constraints were further fused with an INS using the IEKF. Simulation and experimental results indicated that the performance of the trifocal tensor constraints exceeded the three-view geometry constraints. This paper applied the IEKF to analyse the navigation system. As we know, the UKF uses a selected set of points to more accurately map the probability distribution of the measurement model than the linearization of the IEKF, leading to faster convergence from inaccurate initial conditions in estimation problems, so using the UKF method to analyse the performance of the navigaiton might be our future work.
\section{Acknowledgment}
The authors are grateful to the editors Zeynep Celik--Butler and Amy DiMaria and anonymous reviewers for their useful suggestions and detailed comments.


%

\appendices
\section{Proof of Equation (3)}

Proof:
From the notation definition of eq.(1) and eq.(2), we know that  $ \widehat{\bm{x}}=[x^{1},x^{2},x^{3}]^{T},\bm{A}=[\bm{a}_{1},\bm{a}_{2},\bm{a}_{3}]=\bm{C}_{c_{1}}^{c_{2}},\bm{a}_{4}=\bm{T}_{12}^{c_{2}},\bm{B}=[\bm{b}_{1},\bm{b}_{2},\bm{b}_{3}]=\bm{C}_{c_{1}}^{c_{3}},\bm{b}_{4}=\bm{T}_{13}^{c_{3}}$, and $ \bm{T}_{i}=\bm{a}_{i}\bm{b}_{4}^{T}-\bm{a}_{4}\bm{b}_{i}^{T}$, then the middle term of eq.(2) can be written as
\begin{equation}
\begin{aligned}
 &\sum_{i}x^{i}\bm{T}_{i}=x^{1}\bm{T}_{1}+x^{2}\bm{T}_{2}+x^{3}\bm{T}_{3}\\
& =x^{1}(\bm{a}_{1}\bm{b}_{4}^{T}-\bm{a}_{4}\bm{b}_{1}^{T})+x^{2}(\bm{a}_{2}\bm{b}_{4}^{T}-\bm{a}_{4}\bm{b}_{2}^{T})+x^{3}(\bm{a}_{3}\bm{b}_{4}^{T}-\bm{a}_{4}\bm{b}_{3}^{T})\\
& =(x^{1}\bm{a}_{1}+x^{2}\bm{a}_{2}+x^{3}\bm{a}_{3})\bm{b}_{4}^{T}-\bm{a}_{4}(x^{1}\bm{b}_{1}^{T}+x^{2}\bm{b}_{2}^{T}+x^{3}\bm{b}_{3}^{T})\\
& =\bm{A}\widehat{\bm{x}}\bm{b}_{4}^{T}-\bm{a}_{4}\widehat{\bm{x}}\bm{B}^{T}
 =\bm{C}_{c_{1}}^{c_{2}}\widehat{\bm{x}}\bm{T}_{13}^{c_{3}T}-\bm{T}_{12}^{c_{2}}\widehat{\bm{x}}\bm{C}_{c_{3}}^{c_{1}}\\
& =\bm{C}_{c_{1}}^{c_{2}}\widehat{\bm{x}}(\bm{T}_{12}^{c_{3}T}+\bm{T}_{23}^{c_{3}T})-\bm{T}_{12}^{c_{2}}\widehat{\bm{x}}\bm{C}_{c_{3}}^{c_{1}} \\
&=\bm{C}_{c_{1}}^{c_{2}}\widehat{\bm{x}}\bm{T}_{23}^{c_{3}T}+(\bm{C}_{c_{1}}^{c_{2}}\widehat{\bm{x}}\bm{T}_{12}^{c_{3}T}-\bm{T}_{12}^{c_{2}}\widehat{\bm{x}}\bm{C}_{c_{3}}^{c_{1}})\\
\end{aligned}
\end{equation}
Substituting eq.(22) into eq.(2) yields
\begin{equation}
\begin{aligned}
& \bm{0}=[\widehat{\bm{x}}^{'}\times](\sum_{i=1}^{3}\widehat{x}^{i}\bm{T}_{i})[\widehat{\bm{x}}^{''}\times]\\
& =[\widehat{\bm{x}}^{'}\times](\bm{C}_{c_{1}}^{c_{2}}\widehat{x}\bm{T}_{23}^{c_{3}T}+(\bm{C}_{c_{1}}^{c_{2}}\widehat{x}\bm{T}_{12}^{c_{3}T}-\bm{T}_{12}^{c_{2}}\widehat{x}\bm{C}_{c_{3}}^{c_{1}}))[\widehat{\bm{x}}^{''}\times]\\
& =[\widehat{\bm{x}}^{'}\times](\bm{C}_{c_{1}}^{c_{2}}\widehat{x}\bm{T}_{23}^{c_{3}T})[\widehat{\bm{x}}^{''}\times]+[\widehat{\bm{x}}^{'}\times](\bm{C}_{c_{1}}^{c_{2}}\widehat{x}\bm{T}_{12}^{c_{3}T}-\bm{T}_{12}^{c_{2}}\widehat{\bm{x}}^{T}\bm{C}_{c_{3}}^{c_{1}})[\widehat{\bm{x}}^{''}\times]\\
& =(\bm{C}_{n_{2}}^{c_{2}}[\widehat{\bm{x}}^{'n_{2}}\times]\bm{C}_{c_{2}}^{n_{2}})(\bm{C}_{c_{1}}^{c_{2}}(\bm{C}_{n_{2}}^{c_{1}}\widehat{\bm{x}}^{n_{2}})(\bm{C}_{n_{2}}^{c_{3}}\bm{T}_{23}^{n_{2}})^{T})(\bm{C}_{n_{2}}^{c_{3}}[\widehat{\bm{x}}^{''n_{2}}\times]\bm{C}_{c_{3}}^{n_{2}})\\
& +(\bm{C}_{n_{2}}^{c_{2}}[\widehat{\bm{x}}^{'n_{2}}\times]\bm{C}_{c_{2}}^{n_{2}})(\bm{C}_{c_{1}}^{c_{2}}(\bm{C}_{n_{2}}^{c_{1}}\widehat{\bm{x}}^{n_{2}})(\bm{C}_{n_{2}}^{c_{3}}\bm{T}_{12}^{n_{2}})^{T}\\
&-(\bm{C}_{n_{2}}^{c_{2}}\bm{T}_{12}^{n_{2}})(\bm{C}_{n_{2}}^{c_{1}}\widehat{\bm{x}}^{n_{2}})^{T}\bm{C}_{c_{3}}^{c_{1}})(\bm{C}_{n_{2}}^{c_{3}}\bm{T}_{23}^{n_{2}})^{T})(\bm{C}_{n_{2}}^{c_{3}}[\widehat{\bm{x}}^{''n_{2}}\times]\bm{C}_{c_{3}}^{n_{2}})\\
& =\bm{C}_{n_{2}}^{c_{2}}[\widehat{\bm{x}}^{'n_{2}}\times]\widehat{\bm{x}}^{n_{2}}\bm{T}_{23}^{n_{2}T}[\widehat{\bm{x}}^{''n_{2}}\times]\bm{C}_{c_{3}}^{n_{2}}\\
&+\bm{C}_{n_{2}}^{c_{2}}[\widehat{\bm{x}}^{'n_{2}}\times](\widehat{\bm{x}}^{n_{2}}\bm{T}_{12}^{n_{2}T}-\bm{T}_{12}^{n_{2}}\widehat{\bm{x}}^{n_{2}T})[\widehat{\bm{x}}^{''n_{2}}\times]\bm{C}_{c_{3}}^{n_{2}}\\
& =\bm{C}_{n_{2}}^{c_{2}}([\widehat{\bm{x}}^{'n_{2}}\times]\widehat{\bm{x}}^{n_{2}}\bm{T}_{23}^{nT}[\widehat{\bm{x}}^{''n_{2}}\times]\\
&+[\widehat{\bm{x}}^{'n_{2}}\times](\widehat{\bm{x}}^{n_{2}}\bm{T}_{12}^{nT}-\bm{T}_{12}^{n}\bm{x}^{n_{2}T})[\widehat{\bm{x}}^{''n_{2}}\times])\bm{C}_{c_{3}}^{n_{2}}\\
& =\bm{C}_{n_{2}}^{c_{2}}([\widehat{\bm{x}}^{'n_{2}}\times]\widehat{\bm{x}}^{n_{2}}\bm{T}_{23}^{T}[\widehat{\bm{x}}^{''n_{2}}\times]-[\widehat{\bm{x}}^{'n_{2}}\times][(\widehat{\bm{x}}^{n_{2}}\times\bm{T}_{12}^{n})\times][\widehat{\bm{x}}^{''n_{2}}\times])\bm{C}_{c_{3}}^{n_{2}}\\
\end{aligned}
\end{equation}

Since the rotation matrixs are both invertible, so pre-multiply $\bm{C}_{n_{2}}^{c_{2}}$ and post-multiply $\bm{C}_{c_{3}}^{n_{2}}$. Here we assume that the navigation system is rotated very slowly and can be neglected, so we define the navigation system as $ n$, then eq.(23) can be simplified as
 \begin{equation}
   [\widehat{\bm{x}}^{'n}\times]\widehat{\bm{x}}^{n}\bm{T}_{23}^{T}[\widehat{\bm{x}}^{''n}\times]-[\widehat{\bm{x}}^{'n}\times][(\widehat{\bm{x}}^{n}\times\bm{T}_{12}^{n})\times][\widehat{\bm{x}}^{''n}\times]=\bm{0}_{3\times3}
\end{equation}  \par
Proved.

\section{Proof of Lemma 1}

Define matrix $\bm{M}=[\bm{\widehat{x}}^{'n}\times]\bm{\widehat{x}}^{n}\bm{T}_{23}^{nT}[\bm{\widehat{x}}^{''n}\times]-[\bm{\widehat{x}}^{'n}\times][(\bm{\widehat{x}}^{n}\times\bm{T}_{12}^{n})\times][\bm{\widehat{x}}^{''n}\times] $, and the elements of matrix  $\bm{M}$ are
 \begin{equation}
 \bm{M}=
  \begin{bmatrix}
 m_{11} & m_{12} & m_{13}\\
 m_{21} & m_{22} & m_{23}\\
 m_{31} & m_{32} & m_{33}\\
 \end{bmatrix}\\
 \end{equation}
where $ m_{ij}$ represents the row $i$ and column $j$ element of $ \bm{M}$, and \\
\begin{equation}
\begin{aligned}
m_{ij}=&-(\bm{\widehat{x}}^{'n}\times\bm{\widehat{x}}^{nT})\bm{E}_{ij}(\bm{\widehat{x}}^{''n}\times\bm{T}_{23}^{n})-(\bm{\widehat{x}}^{'n}\times\bm{\widehat{x}}^{n})^{T}\bm{E}_{ij}(\bm{\widehat{x}}^{''n}\\
&\times\bm{T}_{12}^{n})+(\bm{\widehat{x}}^{'n}\times\bm{T}_{12}^{n})^{T}\bm{E}_{ij}(\bm{\widehat{x}}^{''n}\times\bm{\widehat{x}}^{n})
 \end{aligned}
\end{equation}
with $\bm{E}_{ij}$ represent a $3\times3 $ dimension matrix,with the $ i$ row and the $j$ column element as one and others as zeros.\par
Next we investigate the relationship between the trifocal tensor constraints and the three-view geometry
constraints proposed by Indelman$^{[16]}$.
Adding the elements $m_{11},m_{22},m_{33}$ of $ \bm{M}$ matrix together, we have
\begin{equation}
\begin{aligned}
 m_{11}+m_{22}+m_{33}=&-(\widehat{\bm{x}}^{'n}\times\widehat{\bm{x}}^{n})^{T}(\widehat{\bm{x}}^{''n}\times\bm{T}_{23}^{n})\\
 &+(\widehat{\bm{x}}^{n}\times\bm{T}_{12}^{n})^{T}(\widehat{\bm{x}}^{''n}\times\widehat{\bm{x}}^{'n})=\bm{0}
 \end{aligned}
\end{equation}

Rearranging (27) yields
\begin{equation}
 (\widehat{\bm{x}}^{'n}\times\widehat{\bm{x}}^{n})^{T}(\widehat{\bm{x}}^{''n}\times\bm{T}_{23}^{n})=(\widehat{\bm{x}}^{n}\times\bm{T}_{12}^{n})^{T}(\widehat{\bm{x}}^{''n}\times\widehat{\bm{x}}^{'n})
\end{equation}

Equation (28) is just the three-view geometry constraints proposed by Indelman$^{[16]}$. On the other hand, the trifocal tensor contains four linearly independent constraints, while the three-view geometry constraints are less than four, so the trifocal tensor constraints are the sufficient conditions of the three-view geometry constraints. \par
Proved.


\ifCLASSOPTIONcaptionsoff
  \newpage
\fi

\end{document}